\documentclass[10pt,conference]{IEEEtran}

\usepackage{cite}

\usepackage{xcolor}
\newif\ifshowrev
\showrevfalse
\ifshowrev
  \newcommand{\rev}[1]{{\color{blue}#1}}
  \newcommand{\todorev}[1]{{\color{orange}\textbf{[PENDING: #1]}}}
\else
  \newcommand{\rev}[1]{#1}
  \newcommand{\todorev}[1]{{\color{red}\textbf{[??]}}}
\fi

\ifCLASSINFOpdf
   \usepackage[pdftex]{graphicx}
   \graphicspath{{figs/}}
   \DeclareGraphicsExtensions{.pdf,.jpeg,.png}
\else
   \usepackage[dvips]{graphicx}
   \graphicspath{{../figs/}}
   \DeclareGraphicsExtensions{.eps}
\fi
\usepackage[cmex10]{amsmath}
\usepackage{amsthm}

\usepackage{algorithmic}

\usepackage{array}

\ifCLASSOPTIONcompsoc
  \usepackage[caption=false,font=normalsize,labelfont=sf,textfont=sf]{subfig}
\else
  \usepackage[caption=false,font=footnotesize]{subfig}
\fi
\usepackage{url}
\usepackage{lipsum}
\usepackage{color}
\usepackage{xcolor}
\usepackage{multirow}
\usepackage[table]{xcolor}
\usepackage{booktabs} 
\usepackage{amsfonts}

\newif\iffinal
\finalfalse

\iffinal
\else
\usepackage[switch]{lineno}
\fi

\begin{document}
%
\title{Forget or Fine-tune? A Comparative Study of Machine Unlearning Strategies for Noisy Label Correction}


\iffinal



%

\else
\author{
\IEEEauthorblockN{João Lucas Pinto de Santana}
\IEEEauthorblockA{Department of Computing\\
Universidade Federal Rural de
Pernambuco,  Brazil\\
Email: joao.lpsantana@ufrpe.br}
\and
\IEEEauthorblockN{Filipe R. Cordeiro}
\IEEEauthorblockA{Department of Computing\\
Universidade Federal Rural de
Pernambuco, Brazil\\
Email: filipe.rolim@ufrpe.br}
}
\fi

\maketitle

\begin{abstract}
Noisy labels remain a critical challenge for training deep neural networks, since memorizing incorrect labels degrades generalization. Once noisy samples are identified after training, the standard solution is to retrain the model from scratch on the cleaned dataset, which is increasingly expensive as datasets and models grow. Machine Unlearning (MU) has recently emerged as a computationally efficient alternative, but the relative effectiveness of different MU strategies for noisy-label correction remains poorly understood. In this work, we conduct a comparative empirical study of five MU methods (NegGrad, Fine-Tuning (FT), Random Labeling (RL), SalUn, and MUNBa) across symmetric, asymmetric, instance-dependent, and open-set noise on CIFAR-10, CIFAR-100, and the real-world noisy dataset Food-101N. \rev{Our central finding is that the appropriate unlearning strategy is conditioned on the noise structure.} Simple FT is a strong baseline across most closed-set scenarios; RL and SalUn are the most consistently robust methods and, under instance-dependent noise, approach retraining accuracy at a fraction of the computational cost; MUNBa shows advantages mainly under extreme symmetric noise. \rev{Under open-set noise, in contrast, we show that retraining on the cleaned subset degrades accuracy relative to the noisy baseline, so approximating the retrained model is not an adequate objective in this regime.} On Food-101N, all MU methods remain competitive and achieve accuracies close to retraining despite reducing runtime by an order of magnitude. These findings provide practical guidelines for selecting MU strategies for post-training noisy-label correction.
\end{abstract}


\IEEEpeerreviewmaketitle

\section{Introduction}
\label{sec:introduction}

Deep neural networks have demonstrated remarkable performance across computer vision applications~\cite{esteva2021deep}, but robust generalization often depends on the availability of high-quality, accurately annotated datasets~\cite{bansal2022systematic}. In challenging domains such as medical imaging, labeling may be prone to ambiguity due to inter-expert variability, introducing noisy labels into the training data and significantly impairing the performance and generalization of trained models~\cite{frenay_classification_2014}. 

Traditional strategies to address label noise typically involve robust training techniques such as noise-tolerant loss functions, sample filtering, and label correction methods~\cite{carneiro2024machine, cordeiro2025anne}. Although existing training strategies can reduce the impact of noisy labels during training, once the model is trained and noisy samples are identified in a post-training scenario, it is necessary to retrain the model to benefit from the curated samples. In the literature, well-known datasets such as COCO~\cite{coco} and DDSM~\cite{ddsm} have been curated after publication, yielding improved results with models retrained on the curated datasets~\cite{deng2024coconut, cbis_ddsm}. In this work, we focus on the scenario where a model trained on a noisy dataset is available, and noisy samples are identified in the original dataset via automatic filtering~\cite{northcutt2021confident}, manual curation or label updates. The common approach of retraining the model from scratch on the cleaned dataset, although effective, is computationally expensive and impractical for large-scale datasets and large models.

\begin{figure}[htp]
    \centering
    \includegraphics[width=0.84\columnwidth]{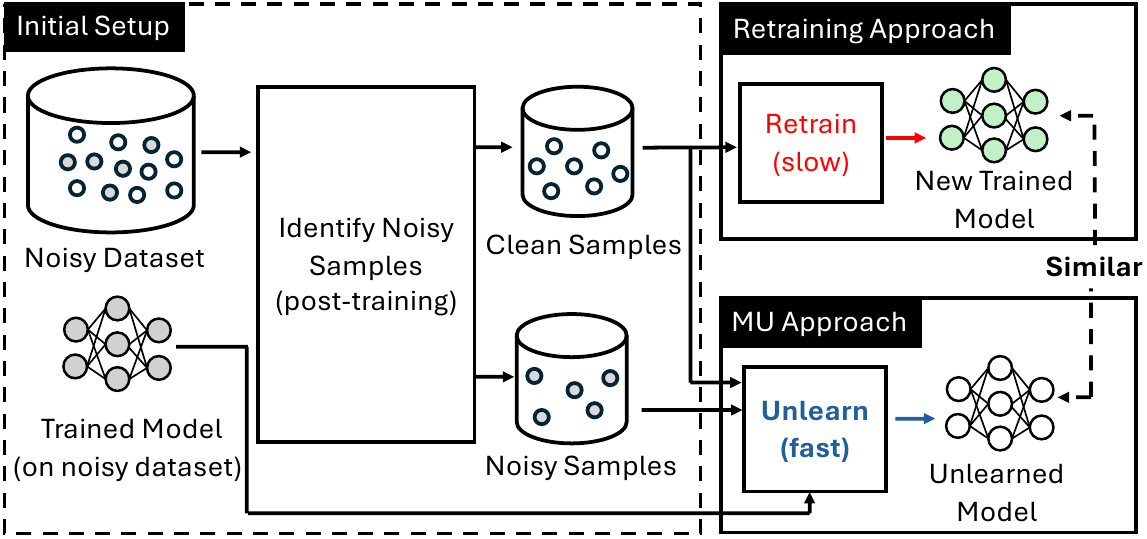}
    \caption{Overview of the retraining and machine unlearning pipelines for correcting a model trained on a noisy dataset, after post-training identification of the noisy samples.}
    \label{fig:setup}
\end{figure}

Machine Unlearning (MU) has emerged as \rev{an} alternative, originally devised to remove sensitive or private data from trained models without necessitating full retraining~\cite{cao2015towards, golatkar2020eternal}. Machine unlearning eliminates the influence of specific data subsets, often requiring only a few epochs of training instead of the hundreds of traditional retraining. 
Recent works have begun to explore MU for noisy label correction via gradient-based unlearning~\cite{sugiura2024removing} and activation projection~\cite{kodge2025sap}.
Figure~\ref{fig:setup} illustrates the scenario considered in this work, where a model trained on a noisy dataset is corrected post-hoc after noisy samples are identified.

Although prior work establishes that MU is applicable to noisy label correction, two questions of direct practical impact remain open. First, the relative effectiveness of different MU strategies, from naive gradient ascent (NegGrad)~\cite{golatkar2020eternal} and simple Fine-Tuning~\cite{warnecke2021machine}, through Random Labeling~\cite{graves2021amnesiac}, to more elaborate methods such as Saliency Unlearning (SalUn)~\cite{fan2023salun} and the bargaining-based MUNBa~\cite{wu2025munba} has not been systematically compared in the noisy label context. Second, since label noise comes in qualitatively different structures (symmetric, asymmetric, instance-dependent, open-set), it is unclear whether the choice of MU method should depend on the noise type, especially since memorization makes unlearning harder~\cite{zhao2024makes} and different noise structures induce different memorization patterns. To address these gaps, we evaluate these five methods on CIFAR-10 and CIFAR-100 with symmetric, asymmetric, instance-dependent and open-set noise, and on Food-101N as a real-world noisy dataset. The main contributions of this work are:

\begin{itemize}
    \item \rev{We present a comparative empirical study of five MU strategies for post-training noisy label correction, over four synthetic noise types and one real-world noisy dataset. Our central finding is that the preferred MU strategy is conditioned on the noise structure: relabeling-based methods (RL, SalUn) are preferable under closed-set and instance-dependent noise, retain-set fine-tuning suffices under asymmetric noise, and balanced forget--retain objectives (MUNBa) become advantageous mainly under  extreme symmetric noise.}
    \item \rev{We show that, under open-set noise, retraining on the cleaned subset degrades test accuracy relative to the model trained on the noisy data: approximating the retrained model is not an adequate objective in this regime, and the decision of whether to unlearn should also be conditioned on the noise structure.}
    \item We find that under instance-dependent noise, relabeling-based unlearning (RL, SalUn) matches full retraining within a few percentage points at roughly twenty times lower cost\rev{, a benefit that persists even under partial identification of the noisy samples}.
\end{itemize}

\section{Related Work}
\label{sec:relatedwork}

Machine Unlearning (MU) is an emerging paradigm designed to remove the influence of specific training samples from a trained model without complete retraining, formalized by Cao and Yang~\cite{cao2015towards}. 


Graves et al.~\cite{graves2021amnesiac} demonstrate that deleting data from the training set is insufficient, as models can still leak information\rev{, and propose relabeling the forget set with random labels, an approach we adopt as the Random Labeling baseline. Saliency Unlearning (SalUn)~\cite{fan2023salun} selectively adjusts the model's weights based on a saliency map of the data to be forgotten, and MUNBa~\cite{wu2025munba} frames unlearning as a cooperative Nash bargaining game between the forgetting and retention objectives.} Zhao et al.~\cite{zhao2024makes} show that examples more highly memorized by the model become more difficult to unlearn, an effect particularly relevant for noisy labels, which are themselves memorized by the model~\cite{zhang_understanding_2017}.

The use of MU for noisy-label correction has recently attracted attention. Sugiura et al.~\cite{sugiura2024removing} investigate the removal of mislabeled data via gradient-based unlearning, Kodge et al.~\cite{kodge2025sap} propose Scaled Activation Projection (SAP), a corrective MU method that projects out activation directions associated with mislabeled samples, and Ye et al.~\cite{ye2025safe} use target label-noise injection to safely unlearn data without performance degradation. Although these works establish the feasibility of MU for noisy labels, each evaluates a single method under a limited set of noise scenarios. \rev{We note that simple fine-tuning is routinely used as a baseline in the MU literature, including in the SalUn protocol~\cite{fan2023salun}. Our contribution is not the observation that FT is competitive per se, but the characterization of \emph{when} it is and is not, as a function of the noise structure.} To the best of our knowledge, no prior work systematically compares distinct MU strategies across qualitatively different noise structures, which is the gap our study addresses.

\section{Problem Setup}
\label{sec:problemsetup}

\subsection{Label Noise}

We denote the training set by $\mathcal{D}=\{(\mathbf{x}_i, \mathbf{y}_i)\}_{i=1}^{|\mathcal{D}|}$, with  $\mathbf{x}_i \in \mathcal{S} \subset \mathbb{R}^{H \times W \times 3}$ being the $i^{th}$ RGB image of size $H \times W$, and $\mathbf{y}_i \in \{0,1\}^{|\mathcal{Y}|}$ denoting a one-hot vector representing the given label, with $\mathcal{Y} = \{1,...,|\mathcal{Y}|\}$ denoting the set of labels, and $\sum_{c \in \mathcal{Y}} \mathbf{y}_i(c)=1$. 
The hidden true label $\hat{\mathbf{y}}_i$ can differ from the given noisy label  $\mathbf{y}_i$ as a result of the label transition probability represented by 
$p(\mathbf{y}(j) = 1 | \mathbf{x}_i,\hat{\mathbf{y}}_i(c) = 1)=\eta_{jc}(\mathbf{x}_i)$,
where the $j,c\in\mathcal{Y}$ are the classes, $\eta_{jc}(\mathbf{x}_i) \in [0,1]$ the probability of flipping the class $c$ to $j$, and $\sum_{j \in \mathcal{Y}}\eta_{jc}(\mathbf{x}_i)=1$. There are four common types of noise in the literature:  
symmetric~\cite{dividemix}, asymmetric~\cite{patrini2017making}, instance-dependent~\cite{rog} and open-set~\cite{cordeiro2025anne}.
The symmetric noise is a label noise type where the hidden true labels are flipped to a random class
with a fixed probability $\eta$, where the true label is included in the label flipping options, which means that $\eta_{jc}(\mathbf{x}_i)=\frac{\eta}{|\mathcal{Y}|-1}, \forall j \in \mathcal{Y}, \text{ such that } j \neq c$, and $\eta_{cc}(\mathbf{x}_i)=1-\eta$. 
The asymmetric noise has its labels flipped 
between similar-looking object categories~\cite{patrini2017making}, where $\eta_{jc}(\mathbf{x}_i)$ depends only on the classes $j,c\in\mathcal{Y}$, but not on $\mathbf{x}_i$. 
The instance-dependent noise~\cite{rog} is the noisy type where the label flipping depends both on the classes $j,c\in\mathcal{Y}$ and image $\mathbf{x}_i$. Two other noise categories are often considered: closed-set noise and open-set noise. In the closed-set noise, the noisy labels are  within the set of valid classes $\mathcal{Y}$, but may not correspond to the hidden true label $\hat{\mathbf{y}}_i \in \mathcal{Y}$.
Conversely, in open-set noise, some noisy labels correspond to samples whose hidden class does not belong to the known label set $\mathcal{Y}$, i.e., $\hat{y}_i \notin \mathcal{Y}$. In this case, noisy samples are drawn from an unknown distribution $\mathcal{O}$  and assigned arbitrary labels $y_i \in \mathcal{Y}$.

\subsection{Label Noise Unlearning}

Machine unlearning can be formally defined as the task of removing the influence of a data subset $\mathcal{D}_f \subset \mathcal{D}$ from  a previously trained model $\theta_o = \mathcal{A}(\mathcal{D})$, where $\theta_o$  is the set of weights resulting from applying a training algorithm $\mathcal{A}$ to the dataset 
$\mathcal{D}$. In the context of label noise unlearning,  $\mathcal{D}_f=\phi(\mathcal{D})$ is a subset of identified noisy samples obtained by a filtering procedure $\phi(\cdot)$. Traditional retraining approaches retrain the model on the remaining subset  $\mathcal{D}_r = \mathcal{D}-\mathcal{D}_f$, obtaining the retrained model weights $\theta_r = \mathcal{A}(\mathcal{D}_r$), without using any data from $\mathcal{D}_f$.
Given this context, the MU task consists of employing an unlearning algorithm $\mathcal{U}$, which, starting from the trained model $\theta_o$, the subset to be forgotten $\mathcal{D}_f$, and the remaining subset $\mathcal{D}_r$, produces an unlearned model $\theta_u = \mathcal{U}(\theta_o, \mathcal{D}_f, \mathcal{D}_r)$.
It is expected that $\theta_u$ approximates, in terms of output distribution, the ideal retrained model $\theta_r$.
The main challenge is to remove the influence of $\mathcal{D}_f$ at a computational cost significantly lower than full retraining, while preserving performance on the remaining data.

\section{Methodology}
\label{sec:methodology}

\subsection{Datasets}
\label{sec:dataset}

We conduct experiments on the datasets CIFAR-10, CIFAR-100~\cite{cifar}, and Food-101N~\cite{lee2018cleannet}. CIFAR-10 and CIFAR-100 have 50k training and 10k testing images of size $32 \times 32$ pixels, with 10 and 100 balanced classes, respectively. As they originally do not contain label noise, following the literature~\cite{dividemix}, we add the following synthetic noise types: symmetric (with noise rate $\eta \in \{0.2, 0.5, 0.8\}$), asymmetric (using the mapping in~\cite{dividemix, patrini2017making}, with $\eta_{jc}=0.4$). We also evaluate CIFAR-10 and CIFAR-100 with instance dependent noise (IDN), following~\cite{xia2020part}, with noise rates in $\{0.2, 0.3, 0.4, 0.5\}$. We also evaluate combined open-set and closed-set noises, as used in~\cite{evidentialmix}. The combined benchmark is defined by the rate of label noise in the experiment, denoted by $\rho \in \{0.3, 0.6\}$, and the proportion of closed-set noise in the label noise, denoted by $\omega \in \{0.5, 1\}$.
The closed-set noise is simulated by symmetrically shuffling the labels of $\rho\,\omega\times100\%$ of the CIFAR-10 training samples, as in~\cite{dividemix}, while the open-set noise replaces $\rho(1-\omega)\times100\%$ of the training images with CIFAR-100 images assigned random CIFAR-10 labels, as in~\cite{cordeiro2025anne}.

Food-101N~\cite{lee2018cleannet} contains 310,009 training images of food recipes in 101 classes, resized to $256 \times 256$, with an estimated label noise of 20\%; the noisy-sample identification provided with the dataset is used as the forget subset. Testing uses the clean 25K-image test set of the original Food101~\cite{bossard2014food}.

\subsection{Compared Machine Unlearning Methods}
\label{sec:methods}

We compare five Machine Unlearning strategies spanning distinct conceptual paradigms, ordered here from the simplest to the most elaborate. \textbf{NegGrad (Gradient Ascent)}~\cite{golatkar2020eternal} is the most direct unlearning operation: it performs gradient \emph{ascent} on the forget set $\mathcal{D}_f$, directly maximizing the loss on the samples to be forgotten, without using the retain set. \textbf{Fine-Tuning (FT)}~\cite{warnecke2021machine} fine-tunes $\theta_o$ on the clean retain set $\mathcal{D}_r$ only, relying on catastrophic forgetting to erode the influence of the removed samples. \textbf{Random Labeling (RL)}~\cite{graves2021amnesiac} follows the Amnesiac paradigm: each sample in $\mathcal{D}_f$ is relabeled with a label drawn uniformly at random from $\mathcal{Y}$, and the model is fine-tuned on $\mathcal{D}_r \cup \mathcal{D}_f'$ for a few epochs, destroying the memorized noisy associations. \textbf{Saliency Unlearning (SalUn)}~\cite{fan2023salun} computes a saliency map from $\nabla_{\theta} \mathcal{L}(\theta_o; \mathcal{D}_f)$ and uses it to mask which parameters receive updates: gradient ascent on $\mathcal{D}_f$ for the top-$k$ salient weights, descent on $\mathcal{D}_r$ for the remaining ones. \textbf{MUNBa}~\cite{wu2025munba} casts unlearning as a cooperative bargaining game: at each step it computes a Pareto-optimal update direction that balances the forgetting gradient on $\mathcal{D}_f$ against the retention gradient on $\mathcal{D}_r$, explicitly avoiding the instability of unconstrained gradient ascent.

\subsection{Implementation}

For the CIFAR-10 and CIFAR-100 datasets, we used a ResNet-18 architecture~\rev{\cite{he2016deep}}, trained for 200 epochs with a learning rate of 0.02\rev{, using} stochastic gradient descent (SGD) with a momentum of 0.9, a weight decay of 0.0005, and a batch size of 256. Each experiment is run three times with independent random seeds for training and unlearning, and we report the mean and standard deviation across the three runs. We used random cropping and horizontal flipping augmentations for the baseline configuration, trained on the noisy dataset as in~\cite{fan2023salun}\rev{, obtaining the trained model $\theta_o$ used as the starting point for unlearning}. For the Food-101N, we used the ResNet-18 architecture and trained the model for 100 epochs, with a learning rate of 0.02 and a batch size of 64.

To generate the retrained model $\theta_r$, we perform full model retraining on the cleaned subset of data $\mathcal{D}_r$, i.e., the training set excluding the identified noisy samples. For CIFAR-10 and CIFAR-100, where noise is synthetically injected, the noisy samples are known a priori and used to define the forget set $\mathcal{D}_f$. The retraining follows the same hyperparameter configuration as the baseline training, applied exclusively to the remaining clean set $\mathcal{D}_r = \mathcal{D} \setminus \mathcal{D}_f$. For the Food-101N, we retrained using the same parameters as the baseline. We used the given dataset's noisy identification to compose $\mathcal{D}_f$.

For all five MU methods, starting from the fully trained model $\theta_o$ on the original noisy dataset $\mathcal{D}$, the model undergoes 10 epochs of unlearning, preserving the original training augmentations. FT, RL, SalUn and MUNBa use an unlearning rate of 0.013, \rev{following the official SalUn evaluation protocol~\cite{fan2023salun}. For NegGrad, we use a lower rate of $10^{-4}$, following the convention adopted in prior gradient-ascent-based unlearning work~\cite{golatkar2020eternal, fan2023salun, thudi2022unrolling}, chosen to prevent the unbounded ascent objective from diverging within the first few unlearning steps. All values are kept fixed across every dataset and noise configuration, without per-scenario tuning.} For the Food-101N dataset, we unlearn for 10 epochs with an unlearning rate of 0.0013 and a batch size of 16. All experiments run on an NVIDIA RTX 4090 GPU. Training and unlearning durations are measured in minutes and reported using the Run-Time Efficiency (RTE) metric, as proposed in~\cite{fan2023salun}. \rev{The code  and configuration files will be publicly released upon acceptance.}

\rev{\textbf{Evaluation scope.} Following the corrective-unlearning perspective of noisy-label removal~\cite{kodge2025sap}, we evaluate all methods by predictive utility (test accuracy) and computational cost (RTE): in this application, the goal of unlearning is to recover the generalization lost to memorized noisy labels, rather than to provide privacy guarantees about $\mathcal{D}_f$. Privacy-oriented forgetting metrics such as membership inference are thus outside our scope, and our conclusions concern MU methods as noisy-label correction tools.}

\section{Results}
\label{sec:results}

We report test accuracy (Acc) and Run-Time Efficiency (RTE, in minutes) for all evaluated methods; accuracies on CIFAR-10, CIFAR-100, and the open-set benchmark are mean$\pm$std over three independent runs, varying both training and unlearning random seeds. \rev{In all tables, the best MU method is shown in bold together with any MU method within one standard deviation of the best mean. }

Table~\ref{tab:res_cifar} presents results on CIFAR-10 with symmetric (20\%, 50\%, 80\%) and asymmetric (40\%) label noise, and on CIFAR-100 with symmetric noise. As expected, the baseline trained directly on noisy labels deteriorates as the noise rate increases, while retraining on the cleaned dataset consistently recovers most of the lost accuracy.

Among the machine unlearning methods, NegGrad is consistently the weakest approach \rev{in the closed-set settings} and becomes increasingly unstable as the noise level grows, while FT, RL, SalUn, and MUNBa substantially improve over the baseline across all evaluated settings. \rev{FT performs on par with the more elaborate methods in several configurations}, particularly under asymmetric noise, where all non-NegGrad approaches achieve accuracies very close to retraining. \rev{At moderate noise levels (e.g., CIFAR-10 with 20\% symmetric noise), RL, SalUn, and FT overlap within one to two standard deviations, so we refrain from declaring a single best method in those cells.}

Under symmetric noise, the differences between methods become more pronounced as the noise rate increases. While FT remains competitive at moderate noise levels, \rev{MUNBa shows an advantage} only under the most challenging settings: under 80\% symmetric noise on CIFAR-10, MUNBa achieves 79.73\% accuracy, \rev{above the remaining MU methods, with a similar trend on CIFAR-100, where the difference between MUNBa and FT falls within one standard deviation}. More elaborate forgetting-retention balancing mechanisms thus become beneficial \rev{mainly} when the amount of memorized label corruption is extremely high\rev{, whereas simple retain-set fine-tuning is sufficient for correcting most asymmetric noise patterns}.

\begin{table*}[htbp]\centering
\caption{Test accuracy (acc, mean$\pm$std over 3 runs) and RTE (min) on CIFAR-10 (sym 20/50/80, asym 40) and CIFAR-100 (sym 20/50/80). \rev{Best MU in \textbf{bold} (ties within one std of the best also bolded).}}
\label{tab:res_cifar}
\scalebox{0.85}{\setlength{\tabcolsep}{2pt}
\begin{tabular}{l|cc|cc|cc|cc|cc|cc|cc}
\toprule
Dataset $\to$ & \multicolumn{8}{c|}{\textbf{CIFAR-10}} & \multicolumn{6}{c}{\textbf{CIFAR-100}} \\\hline
Noise $\to$ & \multicolumn{6}{c|}{Symmetric} & \multicolumn{2}{c|}{Asym} & \multicolumn{6}{c}{Symmetric} \\\hline
Rate $\to$ & \multicolumn{2}{c|}{20\%} & \multicolumn{2}{c|}{50\%} & \multicolumn{2}{c|}{80\%} & \multicolumn{2}{c|}{40\%} & \multicolumn{2}{c|}{20\%} & \multicolumn{2}{c|}{50\%} & \multicolumn{2}{c}{80\%} \\\hline
Method $\downarrow$ & acc & RTE & acc & RTE & acc & RTE & acc & RTE & acc & RTE & acc & RTE & acc & RTE \\
\midrule
Baseline & $90.81_{\text{\tiny $\pm$0.06}}$ & 32.26 & $85.37_{\text{\tiny $\pm$0.52}}$ & 32.26 & $72.65_{\text{\tiny $\pm$0.41}}$ & 32.16 & $89.36_{\text{\tiny $\pm$0.56}}$ & 32.20 & $62.89_{\text{\tiny $\pm$0.30}}$ & 32.16 & $46.23_{\text{\tiny $\pm$0.15}}$ & 32.16 & $20.85_{\text{\tiny $\pm$0.93}}$ & 32.20 \\
Retrain & $94.52_{\text{\tiny $\pm$0.21}}$ & 24.23 & $92.70_{\text{\tiny $\pm$0.13}}$ & 15.10 & $85.24_{\text{\tiny $\pm$0.29}}$ & 4.03 & $93.27_{\text{\tiny $\pm$0.12}}$ & 18.23 & $74.58_{\text{\tiny $\pm$0.16}}$ & 24.20 & $68.80_{\text{\tiny $\pm$0.15}}$ & 15.07 & $51.48_{\text{\tiny $\pm$0.73}}$ & 6.03 \\
\midrule
NegGrad~\cite{golatkar2020eternal} & $90.57_{\text{\tiny $\pm$1.35}}$ & $0.28$ & $86.61_{\text{\tiny $\pm$0.15}}$ & $0.63$ & $74.42_{\text{\tiny $\pm$0.53}}$ & $0.89$ & $89.53_{\text{\tiny $\pm$0.01}}$ & $0.53$ & $63.77_{\text{\tiny $\pm$0.23}}$ & $0.28$ & $47.02_{\text{\tiny $\pm$1.39}}$ & $0.63$ & $17.25_{\text{\tiny $\pm$2.35}}$ & $0.89$ \\
FT~\cite{warnecke2021machine} & $92.05_{\text{\tiny $\pm$0.18}}$ & $1.15$ & $87.93_{\text{\tiny $\pm$0.42}}$ & $0.65$ & $76.79_{\text{\tiny $\pm$0.89}}$ & $0.25$ & $\mathbf{92.79_{\text{\tiny $\pm$0.25}}}$ & $0.80$ & $67.02_{\text{\tiny $\pm$0.38}}$ & $1.15$ & $61.64_{\text{\tiny $\pm$0.51}}$ & $0.65$ & \rev{$\mathbf{36.48_{\text{\tiny $\pm$1.17}}}$} & $0.25$ \\
RL~\cite{graves2021amnesiac} & \rev{$\mathbf{92.33_{\text{\tiny $\pm$0.16}}}$} & $0.95$ & $\mathbf{88.97_{\text{\tiny $\pm$0.47}}}$ & $0.85$ & $76.07_{\text{\tiny $\pm$1.12}}$ & $0.75$ & $92.39_{\text{\tiny $\pm$0.37}}$ & $0.90$ & $\mathbf{68.32_{\text{\tiny $\pm$0.42}}}$ & $1.30$ & $\mathbf{62.57_{\text{\tiny $\pm$0.18}}}$ & $1.15$ & $32.09_{\text{\tiny $\pm$1.35}}$ & $1.00$ \\
SalUn~\cite{fan2023salun} & $\mathbf{92.41_{\text{\tiny $\pm$0.14}}}$ & 1.42 & \rev{$\mathbf{88.96_{\text{\tiny $\pm$0.43}}}$} & 1.26 & $76.19_{\text{\tiny $\pm$0.85}}$ & 1.11 & \rev{$\mathbf{92.58_{\text{\tiny $\pm$0.30}}}$} & 1.32 & \rev{$\mathbf{68.08_{\text{\tiny $\pm$0.29}}}$} & 1.42 & $62.25_{\text{\tiny $\pm$0.16}}$ & 1.26 & $30.70_{\text{\tiny $\pm$1.46}}$ & 1.11 \\
MUNBa~\cite{wu2025munba} & $91.17_{\text{\tiny $\pm$1.02}}$ & $2.84$ & $87.77_{\text{\tiny $\pm$0.69}}$ & $2.52$ & $\mathbf{79.73_{\text{\tiny $\pm$1.02}}}$ & $2.22$ & $92.19_{\text{\tiny $\pm$0.35}}$ & $2.64$ & $67.52_{\text{\tiny $\pm$1.18}}$ & $2.84$ & $61.82_{\text{\tiny $\pm$0.70}}$ & $2.52$ & $\mathbf{38.26_{\text{\tiny $\pm$1.82}}}$ & $2.22$ \\
\bottomrule\end{tabular}}\end{table*}

Table~\ref{tab:idn} reports results under instance-dependent noise (IDN), a more challenging and realistic noise model in which label corruption depends on image content, and the scenario where machine unlearning exhibits its strongest practical potential. RL and SalUn consistently emerge as the most robust MU methods, with RL often providing the best accuracy--runtime trade-off. The strong performance of relabeling-based approaches suggests that actively disrupting memorized noisy associations is particularly effective when corruption follows structured, content-dependent patterns\rev{: unlike} symmetric noise, IDN introduces systematic mistakes embedded in the learned representation\rev{, and relabeling} strategies appear more capable of removing them than retain-only fine-tuning.

Most importantly, RL and SalUn achieve near-retraining performance at a fraction of the computational cost: on CIFAR-100 with 50\% IDN, RL remains within approximately three percentage points of retraining while reducing runtime from 15.10 to 1.15 minutes. These results identify IDN as the most favorable scenario for applying MU instead of full retraining.

\begin{table*}[htbp]\centering
\caption{Test accuracy (mean$\pm$std, 3 runs) and RTE on instance-dependent CIFAR-10/100, noise 20--50\%. \rev{Best MU in \textbf{bold} (ties within one std of the best also bolded).}}
\label{tab:idn}\scalebox{0.79}{\setlength{\tabcolsep}{1.5pt}
\begin{tabular}{l|cc|cc|cc|cc|cc|cc|cc|cc}\toprule
Dataset $\to$ & \multicolumn{8}{c|}{\textbf{IDN-CIFAR-10}} & \multicolumn{8}{c}{\textbf{IDN-CIFAR-100}} \\\hline
Rate $\to$ & \multicolumn{2}{c|}{20\%} & \multicolumn{2}{c|}{30\%} & \multicolumn{2}{c|}{40\%} & \multicolumn{2}{c|}{50\%} & \multicolumn{2}{c|}{20\%} & \multicolumn{2}{c|}{30\%} & \multicolumn{2}{c|}{40\%} & \multicolumn{2}{c}{50\%} \\\hline
Method $\downarrow$ & acc & RTE & acc & RTE & acc & RTE & acc & RTE & acc & RTE & acc & RTE & acc & RTE & acc & RTE \\\midrule
Baseline & $90.73_{\text{\tiny $\pm$0.09}}$ & 32.18 & $88.83_{\text{\tiny $\pm$0.58}}$ & 32.04 & $87.53_{\text{\tiny $\pm$2.63}}$ & 32.10 & $81.93_{\text{\tiny $\pm$5.44}}$ & 32.07 & $63.29_{\text{\tiny $\pm$0.49}}$ & 32.17 & $56.30_{\text{\tiny $\pm$0.76}}$ & 32.05 & $49.81_{\text{\tiny $\pm$0.48}}$ & 32.06 & $45.40_{\text{\tiny $\pm$1.40}}$ & 32.00 \\
Retrain & $93.59_{\text{\tiny $\pm$1.77}}$ & 23.93 & $92.97_{\text{\tiny $\pm$1.61}}$ & 21.20 & $91.76_{\text{\tiny $\pm$2.53}}$ & 18.10 & $91.05_{\text{\tiny $\pm$2.95}}$ & 15.13 & $72.77_{\text{\tiny $\pm$2.98}}$ & 23.86 & $71.27_{\text{\tiny $\pm$3.25}}$ & 21.10 & $68.96_{\text{\tiny $\pm$3.28}}$ & 18.06 & $67.00_{\text{\tiny $\pm$2.96}}$ & 15.10 \\
\midrule
NegGrad~\cite{golatkar2020eternal} & $90.36_{\text{\tiny $\pm$1.71}}$ & $0.28$ & $89.95_{\text{\tiny $\pm$0.20}}$ & $0.41$ & $89.53_{\text{\tiny $\pm$0.01}}$ & $0.53$ & $86.61_{\text{\tiny $\pm$0.16}}$ & $0.64$ & $63.77_{\text{\tiny $\pm$0.23}}$ & $0.28$ & $56.96_{\text{\tiny $\pm$0.08}}$ & $0.41$ & $50.59_{\text{\tiny $\pm$0.13}}$ & $0.53$ & $47.02_{\text{\tiny $\pm$1.39}}$ & $0.64$ \\
FT~\cite{warnecke2021machine} & $91.84_{\text{\tiny $\pm$0.21}}$ & $1.15$ & $91.07_{\text{\tiny $\pm$0.12}}$ & $0.95$ & \rev{$\mathbf{91.71_{\text{\tiny $\pm$1.64}}}$} & $0.80$ & $88.11_{\text{\tiny $\pm$0.73}}$ & $0.65$ & $65.46_{\text{\tiny $\pm$3.06}}$ & $1.15$ & $61.43_{\text{\tiny $\pm$1.36}}$ & $0.95$ & $62.29_{\text{\tiny $\pm$3.12}}$ & $0.80$ & \rev{$\mathbf{62.62_{\text{\tiny $\pm$1.28}}}$} & $0.65$ \\
RL~\cite{graves2021amnesiac} & $92.28_{\text{\tiny $\pm$0.12}}$ & $0.90$ & $\mathbf{91.64_{\text{\tiny $\pm$0.31}}}$ & $0.90$ & \rev{$\mathbf{91.99_{\text{\tiny $\pm$0.32}}}$} & $0.85$ & $\mathbf{89.28_{\text{\tiny $\pm$1.00}}}$ & $0.85$ & \rev{$\mathbf{67.55_{\text{\tiny $\pm$1.74}}}$} & $1.25$ & \rev{$\mathbf{65.08_{\text{\tiny $\pm$1.20}}}$} & $1.25$ & \rev{$\mathbf{64.72_{\text{\tiny $\pm$0.80}}}$} & $1.20$ & $\mathbf{63.92_{\text{\tiny $\pm$2.22}}}$ & $1.15$ \\
SalUn~\cite{fan2023salun} & $\mathbf{92.44_{\text{\tiny $\pm$0.14}}}$ & 1.42 & \rev{$\mathbf{91.52_{\text{\tiny $\pm$0.31}}}$} & 1.37 & $\mathbf{92.16_{\text{\tiny $\pm$0.48}}}$ & 1.32 & \rev{$\mathbf{89.26_{\text{\tiny $\pm$0.95}}}$} & 1.27 & $\mathbf{67.79_{\text{\tiny $\pm$0.79}}}$ & 1.41 & $\mathbf{64.70_{\text{\tiny $\pm$0.11}}}$ & 1.38 & $\mathbf{64.95_{\text{\tiny $\pm$0.28}}}$ & 1.32 & \rev{$\mathbf{63.37_{\text{\tiny $\pm$1.79}}}$} & 1.27 \\
MUNBa~\cite{wu2025munba} & $90.98_{\text{\tiny $\pm$0.77}}$ & $2.84$ & $89.89_{\text{\tiny $\pm$0.54}}$ & $2.74$ & $91.47_{\text{\tiny $\pm$0.90}}$ & $2.64$ & $88.19_{\text{\tiny $\pm$1.43}}$ & $2.54$ & $65.65_{\text{\tiny $\pm$4.41}}$ & $2.82$ & $62.38_{\text{\tiny $\pm$4.53}}$ & $2.76$ & $62.95_{\text{\tiny $\pm$2.97}}$ & $2.64$ & $61.32_{\text{\tiny $\pm$1.56}}$ & $2.54$ \\
\bottomrule\end{tabular}}\end{table*}

\rev{\subsection{Open-Set Noise}
\label{sec:openset_results}}
Table~\ref{tab:open} reports combined open-set and closed-set noise on CIFAR-10, using the format \emph{closed\%/open\%}.
\rev{We first clarify how the baselines of this benchmark are constructed, since they are not directly comparable to those of Table~\ref{tab:res_cifar}. In the configurations with 0\% closed-set noise (0/30 and 0/60), every CIFAR-10 training image keeps its correct label: the corruption consists exclusively of out-of-distribution CIFAR-100 images inserted with arbitrary CIFAR-10 labels, which do not introduce contradictory supervision for the in-distribution classes. The baseline therefore behaves close to a model trained on clean CIFAR-10 (95.11\%), explaining why the open-set baselines exceed those of Table~\ref{tab:res_cifar}, all of which corrupt genuine CIFAR-10 labels.}

\rev{The open-set regime also exposes a limitation of the standard MU objective: in all open-set configurations, the model trained on the noisy data \emph{outperforms} the retrained one (95.11\% vs.\ 93.77\% at 0/30, 95.11\% vs.\ 90.98\% at 0/60, 89.34\% vs. 75.64\% at 30/30). Discarding the identified open-set samples removes real images that, despite their arbitrary labels, still contribute to the learned representation, while the reduced training set penalizes retraining. Consistently, the best MU method in the pure open-set columns is NegGrad, precisely the one that perturbs the model the least. Under open-set noise, test accuracy thus rewards not unlearning: $\theta_r$ is a poor target, and the decision of whether to unlearn at all, not only the choice of method, must be conditioned on the noise structure. When closed-set noise is mixed in (15/15 and 30/30), disrupting the memorized incorrect in-distribution associations becomes beneficial again, and SalUn and RL are the strongest active approaches, surpassing retraining at 30/30.}

\begin{table}[t]\centering
\caption{Test accuracy (mean$\pm$std, 3 runs) and RTE on CIFAR-10 with combined open-set and closed-set noise (closed\%/open\%). \rev{Best MU in \textbf{bold} (ties within one std of the best also bolded).}}
\label{tab:open}\scalebox{0.82}{\setlength{\tabcolsep}{1.5pt}
\begin{tabular}{l|cc|cc|cc|cc}\toprule
Closed/Open $\to$ & \multicolumn{2}{c|}{\textbf{15\%/15\%}} & \multicolumn{2}{c|}{\textbf{0\%/30\%}} & \multicolumn{2}{c|}{\textbf{30\%/30\%}} & \multicolumn{2}{c}{\textbf{0\%/60\%}} \\
Method $\downarrow$ & acc & RTE & acc & RTE & acc & RTE & acc & RTE \\\midrule
Baseline & $91.27_{\text{\tiny $\pm$0.03}}$ & 32.10 & $95.11_{\text{\tiny $\pm$0.07}}$ & 32.07 & $89.34_{\text{\tiny $\pm$0.44}}$ & 32.03 & $95.11_{\text{\tiny $\pm$0.07}}$ & 32.32 \\
Retrain & $87.86_{\text{\tiny $\pm$0.19}}$ & 21.16 & $93.77_{\text{\tiny $\pm$0.02}}$ & 21.13 & $75.64_{\text{\tiny $\pm$0.74}}$ & 12.03 & $90.98_{\text{\tiny $\pm$0.28}}$ & 12.10 \\
\midrule
NegGrad~\cite{golatkar2020eternal} & \rev{$\mathbf{91.23_{\text{\tiny $\pm$0.14}}}$} & $0.41$ & \rev{$\mathbf{95.03_{\text{\tiny $\pm$0.09}}}$} & $0.41$ & $88.75_{\text{\tiny $\pm$0.52}}$ & $0.73$ & \rev{$\mathbf{95.02_{\text{\tiny $\pm$0.08}}}$} & $0.73$ \\
FT~\cite{warnecke2021machine} & $87.20_{\text{\tiny $\pm$0.99}}$ & $0.95$ & \rev{$93.31_{\text{\tiny $\pm$0.05}}$} & $0.95$ & $81.45_{\text{\tiny $\pm$1.33}}$ & $0.50$ & \rev{$93.16_{\text{\tiny $\pm$0.60}}$} & $0.50$ \\
RL~\cite{graves2021amnesiac} & $90.45_{\text{\tiny $\pm$0.02}}$ & $0.90$ & $90.46_{\text{\tiny $\pm$1.00}}$ & $0.90$ & \rev{$\mathbf{89.20_{\text{\tiny $\pm$0.39}}}$} & $0.80$ & $88.37_{\text{\tiny $\pm$1.48}}$ & $0.80$ \\
SalUn~\cite{fan2023salun} & \rev{$90.67_{\text{\tiny $\pm$0.03}}$} & 1.36 & $91.86_{\text{\tiny $\pm$0.20}}$ & 1.36 & $\mathbf{89.32_{\text{\tiny $\pm$0.32}}}$ & 1.21 & $90.57_{\text{\tiny $\pm$0.89}}$ & 1.21 \\
MUNBa~\cite{wu2025munba} & $88.76_{\text{\tiny $\pm$0.53}}$ & $2.72$ & $92.68_{\text{\tiny $\pm$1.17}}$ & $2.72$ & $84.08_{\text{\tiny $\pm$1.19}}$ & $2.42$ & $92.64_{\text{\tiny $\pm$1.27}}$ & $2.42$ \\
\bottomrule\end{tabular}}\end{table}

\rev{\subsection{Real-World Noise}
\label{sec:realworld}}
Table~\ref{tab:food} reports results on Food-101N, a large-scale real-world noisy dataset\rev{, reported from a single run per method due to its computational cost. We acknowledge this limitation in Section~\ref{sec:limitations} and restrict our conclusions on this dataset to differences that are large relative to the seed-level variability observed on the synthetic benchmarks}. All MU methods substantially outperform the baseline trained directly on the noisy dataset: the strongest, RL, reaches 73.10\% accuracy, only 1.09 percentage points below retraining, while reducing runtime from 422.33 to 41.26 minutes. Similarly, FT, SalUn, and MUNBa achieve accuracies within approximately 1.5 percentage points of RL, indicating that the performance differences among modern MU methods become relatively small under real-world label noise. In practical applications, method selection may thus be guided more by computational constraints than by marginal accuracy differences.

\begin{table}[t]
\centering
\caption{Test accuracy and RTE on Food-101N (real-world noise), \rev{single run per method }. Best MU accuracy in \textbf{bold}.}
\label{tab:food}
\scalebox{0.9}{
\begin{tabular}{l|cc}
\toprule
\textbf{Method} & \textbf{Acc} & \textbf{RTE} \\
\midrule
Baseline & 71.30 & 703.33 \\
Retrain  & 74.19 & 422.33 \\
\midrule
NegGrad~\cite{golatkar2020eternal} & 71.19 & $9.06$ \\
FT~\cite{warnecke2021machine}     & 72.84 & $36.12$ \\
RL~\cite{graves2021amnesiac}      & $\mathbf{73.10}$ & $41.26$ \\
 SalUn~\cite{fan2023salun}   & 72.86 & 45.32 \\
MUNBa~\cite{wu2025munba} & 72.75 & $90.64$ \\
\bottomrule
\end{tabular}
}
\end{table}

\begin{figure}[t]
    \centering
    \includegraphics[width=0.86\columnwidth]{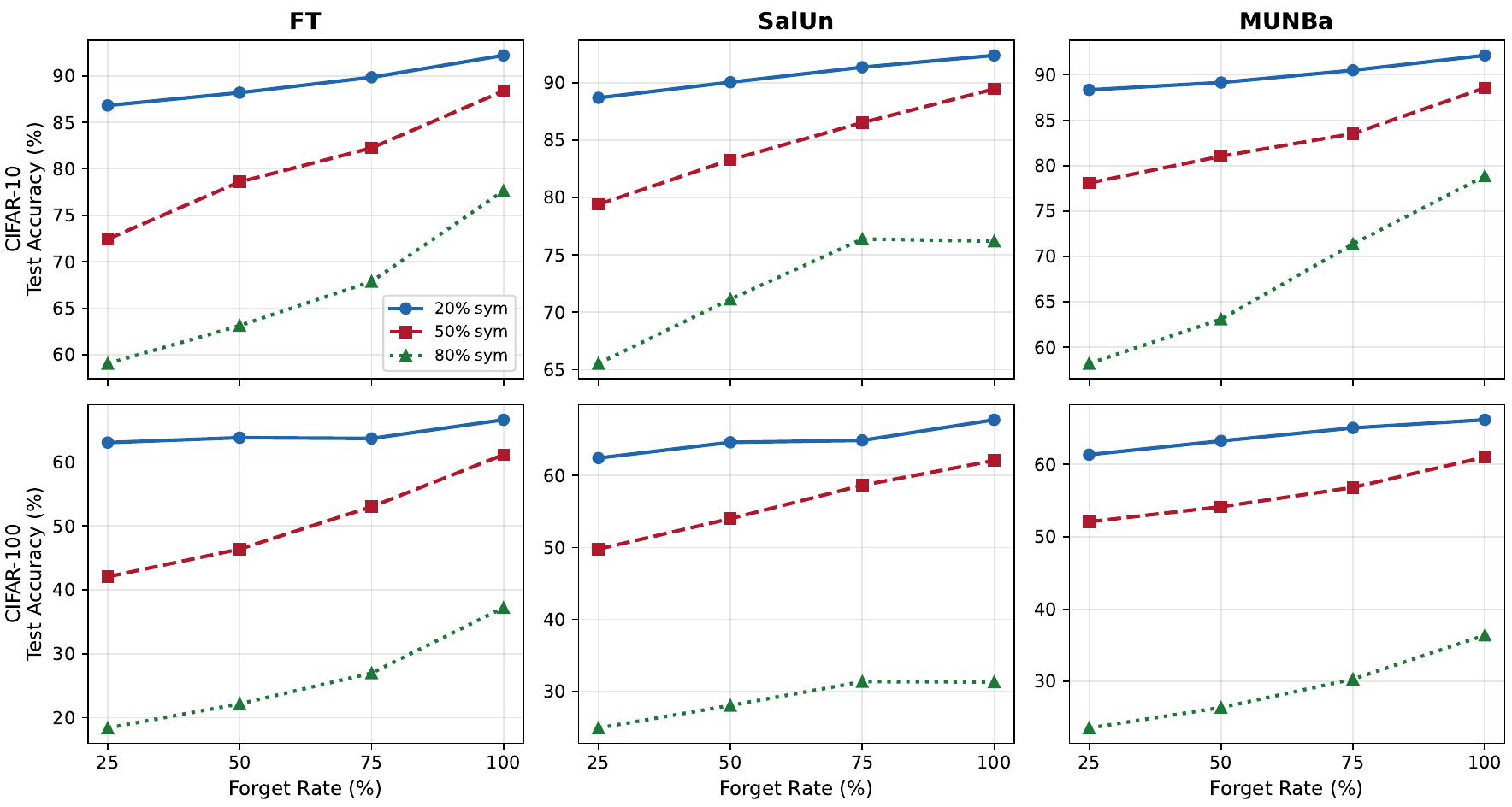}
    \caption{Test accuracy vs.\ forget rate (25--100\% of the identified noisy samples) for FT, SalUn and MUNBa on CIFAR-10 (top) and CIFAR-100 (bottom), with symmetric noise of 20\%, 50\%, and 80\%.}
    \label{fig:forget}
\end{figure}

The experiments above assume that the noisy samples have been correctly identified, but in practice, automatic noise detection is imperfect~\cite{northcutt2021confident}\rev{, producing both false negatives (noisy samples that are missed) and false positives (clean samples wrongly flagged as noisy)}. To assess the impact of \rev{false negatives}, we vary the \emph{forget rate} from 25\% to 100\%, for FT, SalUn and MUNBa on CIFAR-10 and CIFAR-100 with symmetric noise rates of 20\%, 50\%, and 80\% (Figure~\ref{fig:forget}). For all three methods, test accuracy increases monotonically with the forget rate, since the more identified noise is unlearned, the cleaner the effective retain set. Importantly, even partial unlearning yields substantial gains over the baseline. \rev{The complementary failure mode, in which clean samples are wrongly included in $\mathcal{D}_f$, is discussed in Section~\ref{sec:limitations}.}

\rev{\subsection{Discussion and Limitations}
\label{sec:limitations}}
\rev{Two consistent findings emerge from this study: simple Fine-Tuning is a strong MU baseline across most closed-set noise types, and the effectiveness of machine unlearning is strongly influenced by the noise structure, with the open-set regime showing that unlearning toward the retrained model can be counterproductive altogether. Given the runtime reduction relative to retraining, MU is a practical alternative for post-training noisy-label correction within the evaluated scope.}

\rev{That scope is bounded by the following limitations. First, all experiments use a ResNet-18 backbone; although Food-101N (310k images at $256\times256$) provides evidence at a larger data scale, where the runtime gap between unlearning and retraining widens (41 vs.\ 422 minutes), we did not evaluate larger architectures, so the cost argument for large models is extrapolated rather than measured. Second, the number of repetitions (three on the synthetic benchmarks, one on Food-101N) limits the statistical power of per-cell comparisons, which is why we report ties explicitly instead of strict rankings. Third, our evaluation measures MU methods as accuracy-oriented correction tools and does not quantify influence removal through forgetting-specific metrics such as membership inference, so our conclusions concern utility recovery rather than certified forgetting. Fourth, the imperfect-identification analysis covers only missed noisy samples; unlearning falsely flagged clean samples remains untested.}

\section{Conclusion}
\label{sec:conclusion}

In this work, we conducted a \rev{systematic} empirical comparison of five Machine Unlearning strategies for post-training noisy-label correction (NegGrad, FT, RL, SalUn, and MUNBa) on CIFAR-10, CIFAR-100, and Food-101N under symmetric, asymmetric, instance-dependent, open-set, and real-world noise. Our results show that no single MU method dominates all scenarios\rev{: the appropriate strategy is conditioned on the noise structure}. Simple FT constitutes a strong baseline across most \rev{closed-set} settings; RL and SalUn provide the most consistent performance, approaching retraining accuracy at a fraction of the computational cost; the benefits of more elaborate methods such as MUNBa appear primarily under extreme symmetric noise\rev{; and, under open-set noise, retraining on the cleaned subset degrades accuracy relative to the noisy baseline, so the decision of whether to unlearn at all should itself be conditioned on the noise structure}. Overall, in many practical scenarios, simple and efficient approaches recover most of the benefits of full retraining. \rev{Future work includes larger backbones, forgetting-specific metrics such as membership inference, robustness to falsely flagged clean samples, and objectives for the open-set regime that do not rely on approximating the retrained model.}




\bibliographystyle{IEEEtran}
\bibliography{references}
%

\end{document}